# Flood Prediction and Analysis on the Relevance of Features using Explainable Artificial Intelligence

Sai Prasanth Kadiyala, and Wai Lok Woo

Department of Computer and Information Sciences
Northumbria University, England, UK

Email: {sai.kadiyala, wailok.woo}@northumbria.ac.uk

**Abstract:** This paper presents flood prediction models for the state of Kerala in India by analyzing the monthly rainfall data and applying machine learning algorithms including Logistic Regression, K-Nearest Neighbors, Decision Trees, Random Forests, and Support Vector Machine. Although these models have shown high accuracy prediction of the occurrence of flood in a particular year, they do not quantitatively and qualitatively explain the prediction decision. This paper shows how the background features are learned that contributed to the prediction decision and further extended to explain the inner workings with the development of explainable artificial intelligence modules. The obtained results have confirmed the validity of the findings uncovered by the explainer modules basing on the historical flood monthly rainfall data in Kerala.

## 1 Introduction

The increase in floods caused by natural calamities is posing a great danger to the human lives, economy, and environment [1, 2]. Also, the recent increase in the global warming has also led to raise in the frequency of flood occurrence in the entire world [3, 4]. India is among one of the countries being worst hit by floods. Among the states affected Kerala has gained major focus in the year 2108. According to the Central Water Commission (CWC) of India, during August 2018, the state of Kerala (India) witnessed large-scale flooding, which affected millions of people and caused 400 or more deaths [5]. According to the Associated Chambers of Commerce and Industry of India (ASSOCHAM), the damage that Kerala experienced due to this disaster is of worth 15000 to 20000 crores. In the upcoming years it is very much necessary to come up with a solution that can predict the occurrence of the floods in very much advance so that the amount of damage can be reduced.

In the earlier day's traditional models [6] were used to predict events such as storms [7, 8], rainfall [9][10] and other natural calamities. These models showed very good results, but they require many processing units and needed heavy computation which restricts predictions in lesser time [11]. According to the reference [11], the implementation of traditional models requires deep knowledge about parameters in the hydrological study. Apart from the complexity of the models they also showed failure in predicting the floods correctly [12]. Reference [13] presents one such failure of flood prediction that occurred in Queensland, Australia in 2010.Equivalently the numerical prediction models [14] proved unreliable because of their logical mistakes [15].

The advancements in the recent technology have provided a numerous method for developing more precise and accurate applications which can be utilized for better flood predictions. Among such technological advancements, the area of Artificial intelligence and machine learning is one that has gained huge attention due to its low cost of implementation, less amount of coding, low requirements of resources, capability, speed, and accuracy of learning from the earlier events leading to the earlier prediction of an upcoming event [16].

Although the machine learning models have proved to be very beneficial in the technological industry, the insights that the end user has about the internal working of the models to come up with good predictions is very less as these models often are affected by the "black-box" effect [17, 18]. The internal workings of these machine learning models are not easily visible and as the models increase in their complexity, it becomes very difficult to judge if there is any bias and errors in the prediction process. If this issue is not addressed, then the trust among the data hungry models reduces thus giving the user a chance of rejecting them as well in real time scenarios. This is where the Explainable Artificial Intelligence (XAI) comes into place. As far as the authors are concerned, there has been no research undertaken to render the flood prediction model to be more *transparent* and *interpretable* to the users.

The contributions of the paper are to (i) develop of a flood prediction system coupled with the model explain ability, (ii) assess the accuracy of prediction models, and (iii) analyze the validity of the findings uncovered by the explainer modules basing on the historical flood data. To this end, machine learning algorithms such as Logistic Regression, K-Nearest Neighbors (KNN), Decision Trees, and Support Vector Machine have been selected for the development of a flood prediction system and the model explain ability with the help of XAI such as Shapley Adaptive Explanation (SHAP) and Local Interpretable Model-Agnostic Explanation (LIME).

## 2 Background

### *2.1 Area of study*

The area selected for the study is the state of Kerala which is in the southern part of India. In August 2018 the state of Kerala witnessed a very unusual flooding wherein continues rainfall occurred between 8th August to 17th August. Although the state is known to have a lengthy shore and multiple reservoirs, the water could not be stored in the riverbanks leading to a large overflow of water in the Periyar river and Venbanad Lake. As per the Chamber of Commerce India, the cost of state due to the floods is around 2000 crores.

### *2.2 Data*

The dataset for the implementation of the flood prediction model is collected for the state of Kerala for a period of 121 years starting from 1901 to 2021. This dataset [19] consists daily and monthly rainfall, various grouping of monthly rainfall (e.g., one such grouping is January/February, March/April/May, June/July August/Sept, October/November/December) and a column indicating the percentage of flood occurred in that particular year. The predictions are done based on the monthly rainfall occurrence in that particular year. A complete analysis is done on an analysis of the average monthly rainfall from the year 1901 to 2021 is presented in the form of bar graph in Fig. 2 which portrays the months with the highest and lowest rainfall. From Fig. 2, high amount of rainfall concentrates around June and July. This result is obtained as average over the 121 years of rainfall.

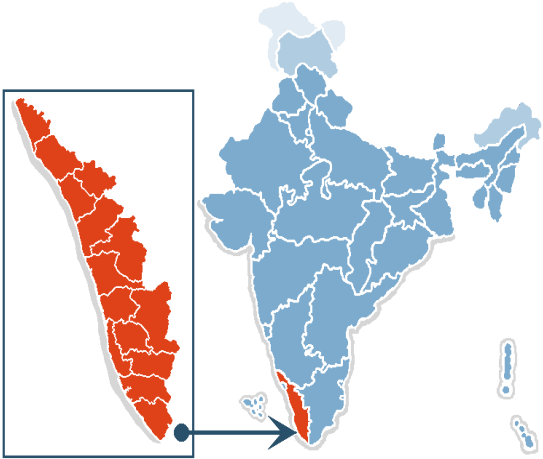 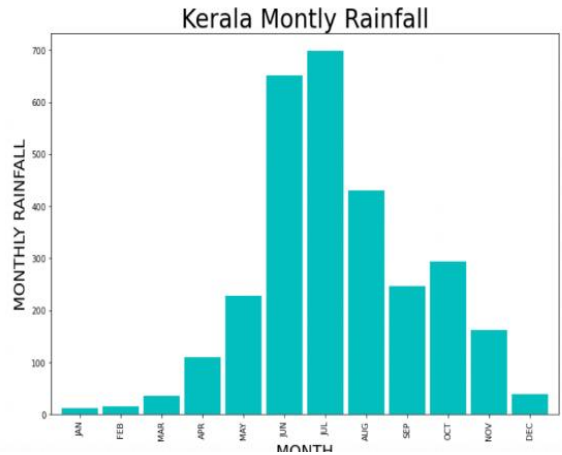

Fig. 1: Map of Kerala     Fig. 2: Monthly rainfall analysis

## 3 Flood Prediction Models

Machine Learning (ML) makes use of mathematical expressions and algorithms for the forecast of floods [20]. Also, the models are data-driven meaning that they work on the previous history of the data and lead to the improvement of the prediction systems through showcase of accurate results in a cost-effective manner [21]. For the successful implementation of a good prediction model, the first step taken into consideration was to pre-process the dataset by checking the existence of any null values and filling up the null values. Secondly the descriptive data present in the dataset is converted into numerical format as the ML models cannot work with categorical form of data directly. Once all the data is encoded to numerical format, the dataset is divided into two separate forms of train dataset (70%) and test dataset (30%). This step is done to verify the if the model is trained by checking the trained model results with the test data results.

### *3.1 K-Nearest Neighbors*

K-Nearest Neighbors (KNN) is a supervised machine learning algorithm that can be used for both classification as well as regression algorithm. But in most of the scenarios, it is used as a classification model [22]. KNN is used for the data that is present in the form of labelled data and classifies the data based on the attributes of its neighbors. The 'K' in the KNN represents the number of nearest neighbors to be taken into consideration for the classification of the new data. The KNN model works based on Euclidean distance formula. As the distance formula, the distance in between of two points in a plane having coordinates $x$, $y$ and $a$, $b$ is obtained by:

$$\sqrt{(x-a)^2 + (y-b)^2} \qquad (1)$$

The algorithm calculates the distance between a particular data that requires to be classified and its nearest neighbors. Basing on the nearest neighbors distance it is classified into that class of data.

*3.2 Logistic regression*

Logistic Regression is a supervised learning model which is used for solving classification problems. It is used when the output is necessary to be present in the 0 or 1, Yes or No, True or False, High or Low. This algorithm works based on the equation below:

$$\log\left[\frac{y}{1-y}\right] = b_0 + b_1 x_1 + b_2 x_2 + \cdots + b_n x_n \qquad (2)$$

*3.3 Decision tree*

Decision tress are supervised learning models with flexibility of the usage in both regression and classification kind of problems. It consists of root nodes, internal nodes, and leaf nodes. The decision tree works by making the question for decision as the root node and based on the question the tree is extended until the least level of entropy is reached [23]. The formula for entropy is given as:

$$\sum_{i=1}^{k} P(value_i) \, log_2(P(value_i)) \qquad (3)$$

where *k* represents the numbers of elements present in the dataset, *P* is the probability of an element. Decision tree gave an accuracy of 75% which is relatively low meaning this would not be the most recommended algorithm for the flood prediction purpose.

*3.4 Support vector machine (SVM)*

SVM is a supervised machine learning model which can be used for classification as well as regression. It uses kernel trick, to transform data and based on these transformations, it forms an optimal boundary (hyperplane) to distinguish the possible outputs. SVM works by classifying the data into various support vectors, then draws an optimal hyperplane in between the vectors keeping in mind that the distance between the hyperplane and the vectors is as far as possible [24].

# 4 Flood Explainers

Though machine have probed to perform better than human beings, one thing that still we tend to doubt is the trust on the solution provided by the machine learning models. Machine learning models perform enormous number of calculations which are hard to understand in the backdrop for making any kind of predictions. This is where the Explainable AI models come into play. XAI makes the interpretation of ml models in a human readable format. XAI comes as a part of Artificial Intelligence (AI) which helping make clear explanation to the human minds [25]. The XAI also helps build trust on the model by letting us know the data that is affecting the prediction results by displaying the features of highest importance in this paper the focus is mainly only the XAI methods LIME and SHAP.

## *4.1 Shapley Additive Explanations (SHAP)*

The SHAP is built on a game theory of Shapley values which was invented by Lloyd in 1952. The theory was invented to answer a situation where, suppose we have a combination of elements forming an item $X$ which when put together produces an output $Y$, what would be the individual contribution that leads to the outcome. SHAP mainly tries to interpret how much each individual feature of contribute to attain the outcome. In the SHAP, additive is defined as follows: Suppose we have a set of input $x$ and a model $f(x)$, and a simplified local input $x'$, an explanatory model $g$ what need to be seen is that if $x'$ is roughly equal to $x$ the $g(x')$ also should be roughly equal to $f(x')$ and $g$ must take the form:

$$g(x') = \varphi_0 + \sum_{i=1}^{N} \varphi_i x'_i \qquad (4)$$

where $\varphi_0$ is the null output or the average output of the model and $\varphi_i$ is how much the feature $i$ changed the model outcome [26].

The problem with computing the Shapley values is having to sample the coalition values for every possible feature permutation, which means that we need to evaluate a model multiple times leading to huge time consumption and difficulty. To overcome this problem Lundberg and Lee came up with the Shapley Kernel which works by approximating Shapley values with less samples. In this paper, SHAP kernel is used to understand each month's contribution in the cause of floods and understand the three months of feature importance responsible for floods.

*4.2 Local Interpretable Model Explanation (LIME)*

Local Interpretable Model Explanation (LIME) checks if the prediction that has been made is close to the expected model results or not. The LIME focuses on local interpretability, which is determined by accessing only one input feature that fits in a line of linearity using a regularization constraint applied to a linear regression model. LIME takes a single data input and checks for all the features in the data that are made responsible for the prediction of the machine learning models and classified them into two categories on either side of a line where the left-hand side of the line shows the features in the dataset that are having negative impact on the model prediction and the right-hand side of the line represents the features that are having a positive impact on the machine learning model predictions. In this way the LIME builds the trust on the results by clearly displaying the interior working of the machine learning models. The mathematical interpretation of the LIME model is as follows:

$$E(x) = \underset{g \in G}{\operatorname{argmin}} L(f, g, \pi_x) + \Omega(g) \tag{5}$$

where *L* is loss, $\Omega$ is penalty for model complexity, *g* is the model, *x* is the data point [27].

## 5 Results, Analysis and Validation

*5.1 Model selection*

All prediction models will be evaluated in terms of the accuracy, precision, recall and F1-score. Table I shows the performance of the prediction based on 70% training and 30% test dataset. It can be noticed from Table I that logistic regression has shown am accuracy of 0.95, recall score of o.95 meaning that there are very less chances of falsely predicting a positive value and the F1 score of o.95 shows that there is a very good balance between the precision and recall scores meaning that the overall performance of Logistic Regression is very good for the prediction of floods. While the other model KNN shows the least accuracy of 0.75 with very less precision, recall and F1-scores of 0.8, 0.66, and 0.72. The remaining two models' decision trees and support vector machine also do not show expected efficiency when we look at their metric scores. From the above study it can be clearly seen that the Logistic Regression has outperformed the remaining three machine learning models, making it the best recommendable machine learning model for the accurate prediction of floods. Hence, it will be used to analyze the relevance of features.

Table I: Prediction results on test dataset

| Model | Accuracy | Precision | Recall | F1-score |
|---|---|---|---|---|
| Logistic Regression | 0.95 | 1.0 | 0.91 | 0.95 |
| KNN | 0.75 | 0.8 | 0.66 | 0.72 |
| Decision Tree | 0.83 | 0.9 | 0.75 | 0.81 |
| SVM | 0.87 | 0.91 | 0.84 | 0.87 |

*5.2 Flood analysis with Explainable AI*

After performing the implementation of the machine learning algorithms by training and testing the data Logistic regression looks to perform the best with an accuracy of 0.95. But to understand the internal working of the logistic regression model it important to open the Blackbox and learn its working with the help of SHAP and LIME.

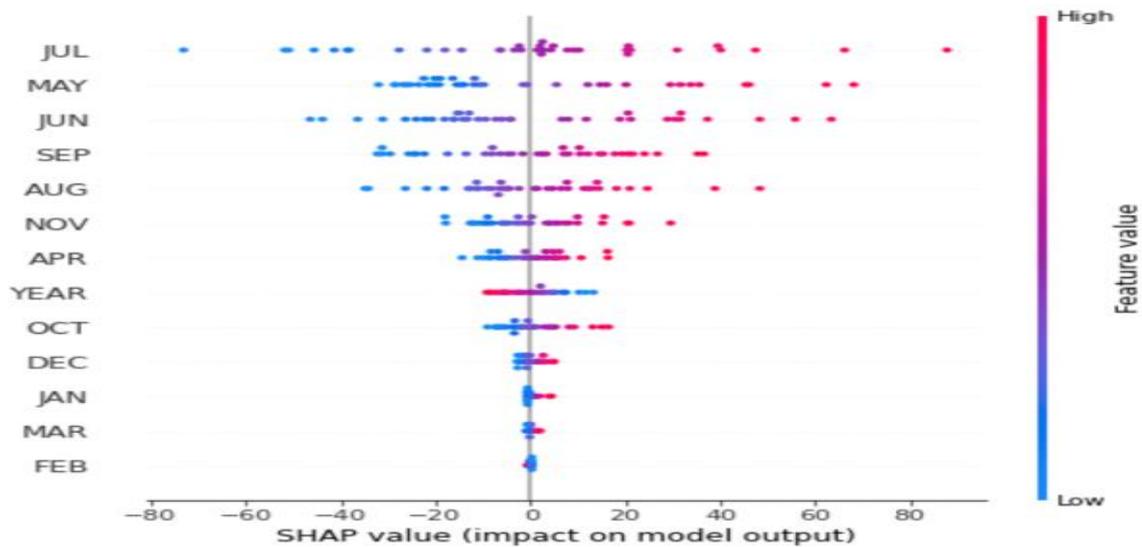

Fig 3: Features contribution to flood and no-flood from 1901 to 2021 by SHAP

**SHAP**: The SHAP has been used to identify the feature importance in the model prediction. This mainly shows the contribution of each month's rainfall data for the machine learning model to predict floods or no floods. From Fig. 3, the SHAP mainly shows the months that have impact on the model output in a descending order where it clearly indicates that the rainfall data of the month July has the highest impact on the model, followed by month May, June, September, and August. The November, April, October contribute to the model at medium level for the prediction and the months with the least impact on the model are January, February, and March. This shows the global analysis of feature contributions from 1901 to 2021 as revealed by SHAP.

**LIME**: After implementing the LIME, the following results have been identified. In 1947, there was a flood in Kerala. LIME output the explanation and analysis in Fig. 4.

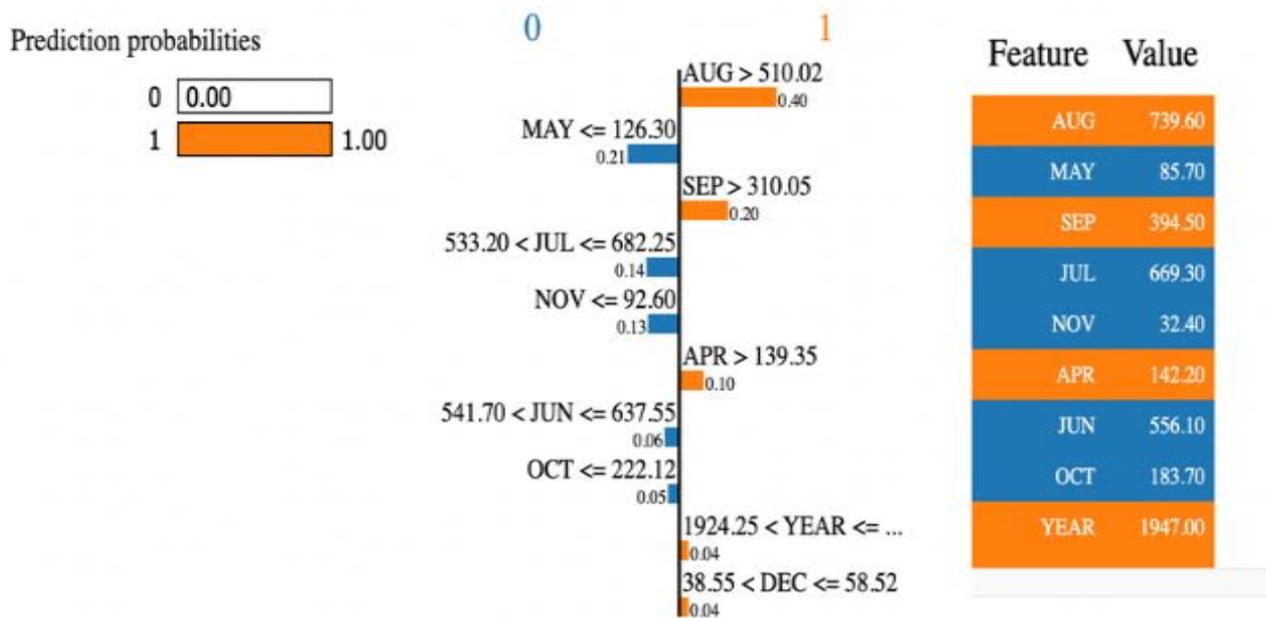

Fig 4: Features contribution to flood and no-flood in 1947 by LIME

Fig. 4 shows that the year 1947 has the complete possibility of flood as the months August contributes 739 meters of rainfall which is greater than the threshold value of 510.02 that has been set by the lime to classify into month that causes flood or not. Similarly, September can be seen as the month with heavy rainfall causing the result to be as flood. The April month contribution also contributed to flood and the December contributed the least to the prediction of floods in that year. It is interesting to note that LIME reveals the local analysis of feature contributions for the particular year.

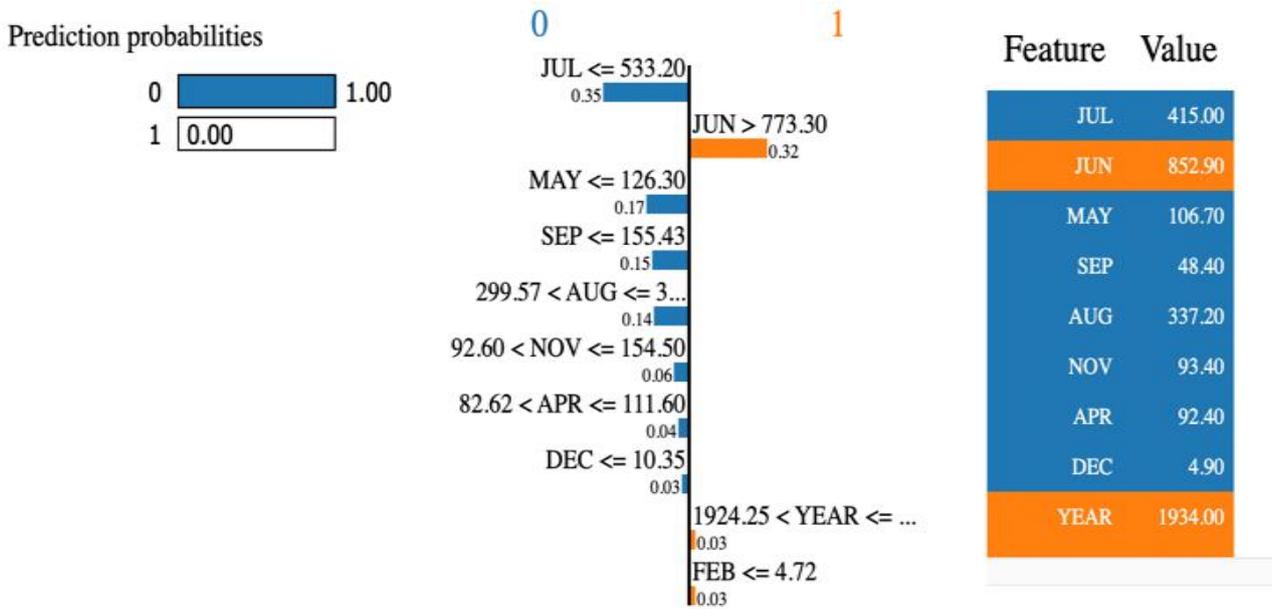

Fig 5: Features contribution to flood and no-flood in 1934 by LIME

Comparing the results that have been identified in LIME with the overall results provided by the SHAP, the months August, September are among the top five months and April contributes at medium for the prediction of the model results in SHAP. The same is seen in the LIME results which show the month August and September contributing the highest to the model output in the year 1947 for a flood occurrence. Fig. 5 depicts the LIME prediction which shows that there is no flood in the 1934. This is validated by the fact that July has the highest contribution to the model results with the month having rainfall of 415.0 which is very less leading to the prediction as no floods in the 1934.Comparing the SHAP and LIME results, SHAP shows that July, June, August, May, and September contribute the highest to the model output and similarly in LIME the same months are responsible for the correct prediction of the output. Comparing the results of the SHAP and LIME, both the models project the months May, June, July, and September as the months with the highest impact on the model prediction.

# 6 Conclusion

The paper has presented the need for a machine learning based prediction model for prediction of floods. Four different machine learning models KNN, Decision Tree, Logistic Regression, and Support Vector Machine have been compared by their metric scores including accuracy, precision, recall and F1-score. Results have shown that Logistic Regression is the best model with highest metric score. The paper has further validated the working of the machine learning model with the Explainable AI models i.e., SHAP and LIME, which respectively reveal the global and local feature contributions to prediction of flood and no-flood. The future direction of the work is to consider deep learning models [28-29] and human-machine interaction [30] where this framework can enable users to find a solution that can help forecast the flooding earlier in the coming years.